\documentclass[11pt]{article}
\let\atsign@
\usepackage{amstex}

\usepackage{times}
\usepackage{mathfont}
\usepackage{graphicx}
\usepackage{cite}
\usepackage{url}
\def\max{\mathop{\rm max}}
\def\gcd{\mathop{\rm gcd}}
\def\evolve{\mathop{\rm evolve}}

\DeclareSymbolFont{AMSb}{U}{msb}{m}{n}
\DeclareSymbolFontAlphabet{\Bbb}{AMSb}

\begin{document}

\title{Searching for Spaceships}

\author{David Eppstein
\thanks{Dept. of Information and Computer Science,
U. of California, Irvine, CA 92697-3425, 
eppstein\atsign ics.uci.edu.}}

\date{}
\maketitle   
 
\begin{abstract}
We describe software that searches for spaceships in Conway's Game of
Life and related two-dimensional cellular automata. Our program searches
through a state space related to the de Bruijn graph of the automaton,
using a method that combines features of breadth first and iterative
deepening search, and includes fast bit-parallel graph reachability
and path enumeration algorithms for finding the successors of each state.
Successful results include a new $2c/7$ spaceship in Life, found by
searching a space with
$2^{126}$ states.
\end{abstract}

\section{Introduction}

\begin{figure}[t]
$$\includegraphics{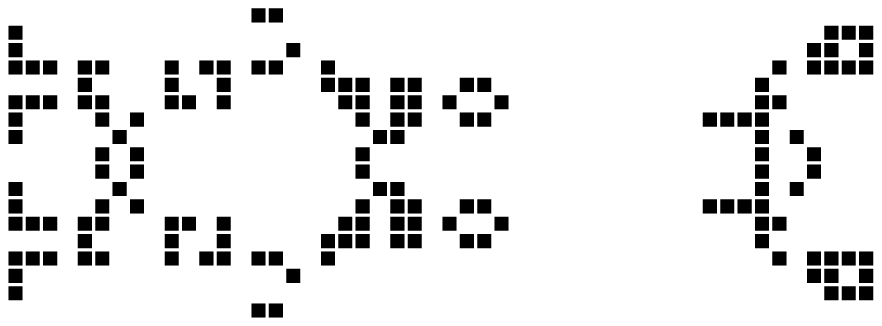}$$
\caption{The ``dragon'' (left) and ``weekender'' (right) spaceships in
Conway's Life (B3/S23).  The dragon moves right one step every six
generations (speed $c/6$) while the weekender moves right two steps
every seven generations (speed $2c/7$).}
\label{fig:weekender}
\end{figure}

\begin{figure}[t]
$$\includegraphics{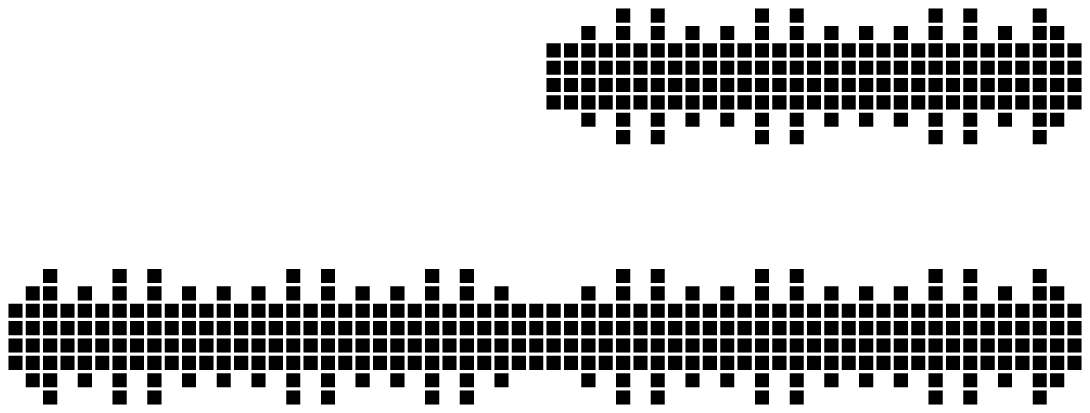}$$
\caption{A $c/7$ spaceship in the ``Diamoeba'' rule (B35678/S5678),
top, and a spacefilling pattern formed by two back-to-back spaceships.}
\label{fig:diamoeba}
\end{figure}

John Conway's Game of Life has fascinated and inspired many enthusiasts,
due to the emergence of complex behavior from a very simple system.
One of the many interesting phenomena in Life is the existence of
gliders and spaceships: small patterns that move across space. When
describing gliders, spaceships, and other early discoveries in Life,
Martin Gardner wrote (in 1970) that spaceships ``are extremely hard to
find''
\cite{Gar-SI-70}.  Very small spaceships can be found by human
experimentation, but finding larger ones requires more
sophisticated methods. Can computer software aid in this search? The
answer is yes -- we describe here a program, {\tt gfind}, that can quickly
find large low-period spaceships in Life and many related cellular
automata.

Among the interesting new patterns found by {\tt gfind}
are the ``weekender'' $2c/7$ spaceship in Conway's Life
(Figure~\ref{fig:weekender}, right), the ``dragon'' $c/6$
Life spaceship found by Paul Tooke (Figure~\ref{fig:weekender}, left),
and a $c/7$ spaceship in the Diamoeba
rule (Figure~\ref{fig:diamoeba}, top). The middle section of the Diamoeba
spaceship simulates a simple one-dimensional parity automaton and can be
extended to arbitrary lengths. David Bell discovered that two back-to-back
copies of these spaceships form a pattern that fills space with
live cells (Figure~\ref{fig:diamoeba}, bottom). 
The existence of
infinite-growth patterns in Diamoeba had previously been posed as
an open problem (with a \$50 bounty) by Dean Hickerson in August 1993, and
was later included in a list of related open problems by Gravner and
Griffeath \cite{GraGri-AAM-98}. Our program has also found new spaceships
in well known rules such as HighLife and Day\&Night as well
as in thousands of unnamed rules.

As well as providing a useful tool for discovering cellular automaton
behavior, our work may be of interest for its use of state space search
techniques. Recently, Buckingham and Callahan \cite{BucCal-EM-98} wrote
``So far, computers have primarily been used to speed up the design
process and fill gaps left by a manual search. Much potential remains
for increasing the level of automation, suggesting that Life may merit
more attention from the computer search community.''  Spaceship
searching provides a search problem with characteristics intriguingly
different from standard test cases such as the 15-puzzle or computer
chess, including a large state space that fluctuates in width instead of
growing exponentially at each level, a tendency for many branches of the
search to lead to dead ends, and the lack of any kind of admissable
estimate for the distance to a goal state.  Therefore, the search
community may benefit from more attention to Life.

The software described here, a database of
spaceships in Life-like automata, and several programs for
related computations can be found online at
\url{http://www.ics.uci.edu/~eppstein/ca/}.

\section{A Brief History of Spaceship Searching}

\begin{figure}[t]
$$\includegraphics{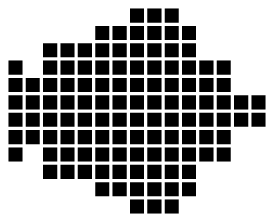}$$
\caption{Large $c/60$ spaceship in rule B36/S035678,
found by brute force search.}
\label{fig:p60}
\end{figure}

According to Berlekamp et al. \cite{BerConGuy-82}, the $c/4$
diagonal glider in Life was first discovered by
simulating the evolution of the R-pentomino, one of only 21 connected
patterns of at most five live cells.  This number is small enough that
the selection of patterns was likely performed by hand.  Gliders can also
be seen in the evolution of random initial conditions in Life as
well as other automata such as B3/S13 \cite{Heu-CS-96}, but this
technique often fails to work in other automata due to the lack of large
enough regions of dead cells for the spaceships to fly through. Soon
after the discovery of the glider, Life's three small $c/2$ orthogonal
spaceships were also discovered.

Probably the first automatic search method developed to look for
interesting patterns in Life and other cellular automata,
and the method most commonly programmed, is
a brute force search that tests patterns of bounded size,
patterns with a bounded number of live cells, or patterns formed out of
a small number of known building blocks.  These tests might be
exhaustive (trying all possible patterns) or they might perform a sequence
of trials on randomly chosen small patterns.  Such methods have found many
interesting oscillators and other patterns in Life, and Bob Wainwright
collected a large list of small spaceships found in this way
for many other cellular automaton rules \cite{Wai-94}.  Currently, it is
possible to try all patterns that fit within rectangles of up to
$7\times 8$ cells (assuming symmetric initial conditions), and this
sort of exhaustive search can sometimes find spaceships as
large as $12\times 15$ (Figure~\ref{fig:p60}). However, brute force
methods have not been able to find spaceships in Life with speeds other
than $c/2$ and $c/4$.

The first use of more sophisticated search techniques came in 1989,
when Dean Hickerson wrote a backtracking search program which he called
{\tt LS}.  For each generation of each cell inside a fixed rectangle,
{\tt LS} stored one of three states: unknown, live, or dead.  {\tt LS}
then attempted to set the state of each unknown cell by examining
neighboring cells in the next and previous generations.  If no unknown
cell's state could be determined, the program performed a depth first
branching step in which it tried both possible states for one of the
cells.  Using this program, Hickerson discovered many patterns including
Life's $c/3$,
$c/4$, and
$2c/5$ orthogonal spaceships.  Hartmut Holzwart used a similar program
to find many variant $c/2$ and $c/3$ spaceships in Life, and related
patterns including the ``spacefiller'' in which four $c/2$ spaceships
stretch the corners of a growing diamond shaped still life. David Bell
reimplemented this method in portable C, and added a fourth ``don't
care'' state; his program, {\tt lifesrc}, is available at
\url{http://www.canb.auug.org.au/~dbell/programs/lifesrc-3.7.tar.gz}.

In 1996, Tim Coe discovered another Life spaceship, moving orthogonally
at speed $c/5$, using a program he called {\tt knight} in the hope that
it could also find ``knightships'' such as those in
Figure~\ref{fig:knights}. {\tt Knight} used breadth first search (with
a fixed amount of depth-first lookahead per node to reduce the space
needs of BFS) on a representation of the problem based on {\em de
Bruijn graphs} (described in Section~\ref{sec:state}). The
search took 38 cpu-weeks of time on a combination of Pentium Pro 133 and
Hypersparc processors. The new search program we describe here can be
viewed as using a similar state space with improved search algorithms and
fast implementation techniques. Another recent program by Keith Amling
also uses a state space very similar to Coe's, with a depth first search
algorithm.

Other techniques for finding
cellular automaton patterns include complementation of
nondeterministic finite automata, by Jean Hardouin-Duparc
\cite{Har-PMUBA-72/73,Har-RAIRO-74};
strong connectivity analysis of
de Bruijn graphs, by Harold McIntosh
\cite{McI-still-88,McI-zoo-88}; randomized hill-climbing methods
for minimizing the number of cells with incorrect evolution, by Paul
Callahan (\url{http://www.cs.jhu.edu/~callahan/stilledit.html});
a backtracking search for still life backgrounds such that an
initial perturbation remains bounded in size as it evolves, by
Dean Hickerson; Gr\"obner basis methods, by John Aspinall;
and a formulation of the search problem as an integer
program, attempted by Richard Schroeppel and later applied
with more success by Robert Bosch \cite{Bos-SR-99}.
However to our knowledge none of these techniques has been used to find
new spaceships.

% Aspinall and Schroeppel: see messages dated 15 Sep 1993

\section{Notation and Classification of Patterns}

\begin{figure}[t]
$$\includegraphics{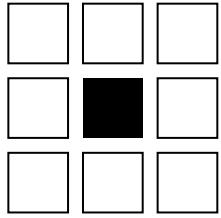}$$
\caption{The eight neighbors in the Moore neighborhood of a cell.}
\label{fig:Moore}
\end{figure}

\begin{figure}[t]
$$\includegraphics{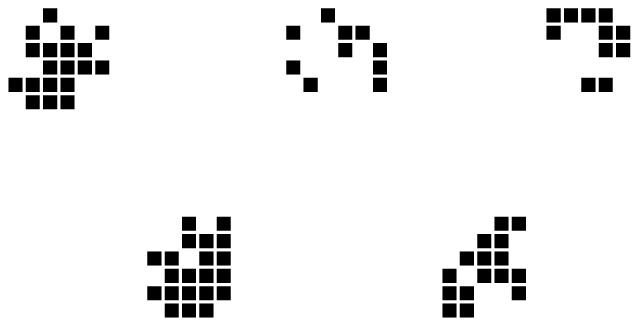}$$
\caption{Slope 2 and slope 3/2 spaceships. Left to right:
(a) B356/S02456, $2c/11$, slope 2.
(b) B3/S01367, $c/13$, slope 2.
(c) B36/S01347, $2c/23$, slope 2.
(d) B34578/S358, $2c/25$, slope 2.
(e) B345/S126, $3c/23$, slope 3/2.}
\label{fig:knights}
\end{figure}

%% insufficiently relevant
%
% \begin{figure}[t]
% $$\includegraphics{b368s12578.eps}$$
% \caption{Period 104 $c/8$ diagonal spaceship gun in rule B368/S12578,
% formed by the interaction of three replicator-based oscillators.}
% \label{fig:b368s12578}
% \end{figure}

\begin{figure}[t]
$$\includegraphics{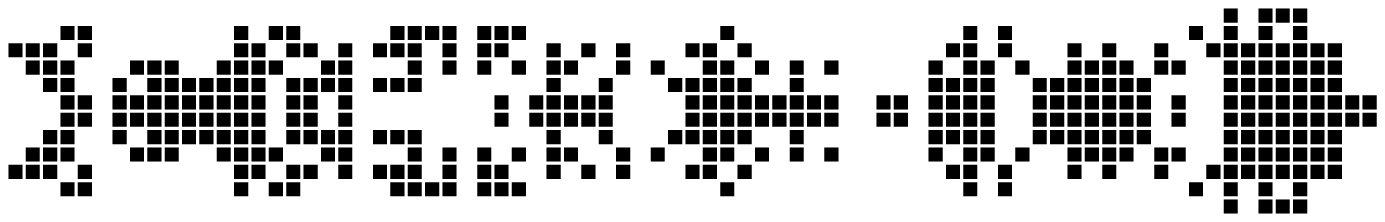}$$
\caption{Long narrow c/6 spaceship in Day\&Night (B3678/S34678).}
\label{fig:dn6}
\end{figure}

\begin{figure}[t]
$$\includegraphics{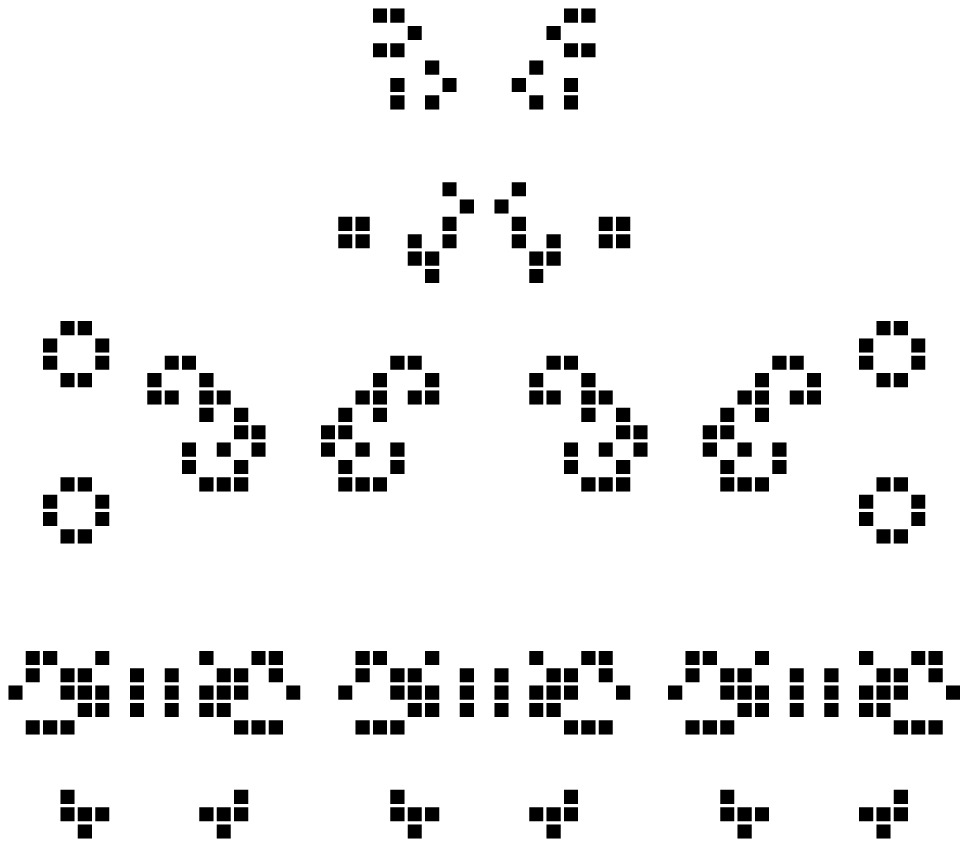}$$
\caption{$9c/28$ spaceship formed from 10 R-pentomino puffers in
B37/S23.}
\label{fig:puddlejump}
\end{figure}

We consider here only {\em outer totalistic} rules, like Life, in which
any cell is either ``live'' or ``dead'' and in which the state of any
cell depends only on its previous state and on the total number of live
neighbors among the eight adjacent cells of the Moore neighborhood
(Figure~\ref{fig:Moore}). For some results on spaceships in cellular
automata with larger neighborhoods, see Evans' thesis \cite{Eva-PhD-96}.
Outer totalistic rules are described with a string of the
form B$x_1x_2\ldots$/S$y_1y_2\ldots$ where the $x_i$ are digits listing
the number of neighbors required for a cell to be born (change from dead
to live) while the $y_i$ list the number of neighbors required for a
cell to survive (stay in the live state after already being live).
For instance, Conway's Life is described in this notation as B3/S23.

A {\em spaceship} is a pattern which repeats itself after some number $p$
of generations, in a different position from where it started.
We call $p$ the {\em period} of the
spaceship. If the pattern moves $x$ units horizontally and $y$ units
vertically every $p$ steps, we say that it has slope $y/x$ and speed
$\max(|x|,|y|)c/p$, where $c$ denotes the maximum speed at which
information can propagate in the automaton (the so-called {\em speed of
light}). Most known spaceships move orthogonally or diagonally, and many
have an axis of symmetry parallel to the direction of motion.  Others,
such as the glider and small $c/2$ spaceships in Life, have {\em
glide-reflect symmetry}: a mirror image of the original pattern appears
in generations $p/2$,
$3p/2$, etc.; spaceships with this type of symmetry must also move
orthogonally or diagonally. However, a few asymmetric spaceships move
along lines of slope 2 or even 3/2 (Figure~\ref{fig:knights}).
According to Berlekamp et al. \cite{BerConGuy-82},
there exist spaceships in Life that move with any given rational slope,
but the argument for the existence of such spaceships is not very explicit
and would lead to extremely large patterns.

Related types of patterns include oscillators (patterns that repeat in
the same position), still lifes (oscillators with
period 1), puffers (patterns which repeat some distance away after a
fixed period, leaving behind a trail of discrete patterns such
as oscillators, still lifes, and spaceships), rakes (spaceship puffers),
guns (oscillators which send out a moving trail of discrete patterns
such as spaceships or rakes), wickstretchers (patterns which leave
behind one or more connected stable or oscillating regions), replicators
(patterns which produce multiple copies of themselves), breeders
(patterns which fill a quadratically-growing area of space with discrete
patterns, for instance replicator puffers or rake guns), and spacefillers
(patterns which fill a quadratically-growing area with one or more
connected patterns).
% Often, patterns of one type can be used as components in a pattern
% of another type (Figures \ref{fig:b368s12578} and~\ref{fig:puddlejump}).
See Paul Callahan's Life Pattern Catalog
(\url{http://www.cs.jhu.edu/~callahan/patterns/contents.html}) for
examples of many of these types of patterns in Life, or
\url{http://www.ics.uci.edu/~eppstein/ca/replicators/}
for a number of replicator-based patterns in other rules.

We can distinguish among several classes of spaceships, according to
the methods that work best for finding them.

\begin{itemize}
\item Spaceships with small size but possibly high period can be found by
brute force search. The
patterns depicted in Figures \ref{fig:p60} and~\ref{fig:knights} fall into
this class, as do Life's glider and c/2 spaceships.

\item Spaceships in which the period and one dimension are small,
while the other dimension may be large, can be found by search
algorithms similar to the ones described in this paper.
The new Life spaceships in Figure~\ref{fig:weekender}
fall into this class.  For a more extreme example of a low period ship
which is long but narrow, see Figure~\ref{fig:dn6}.
``Small'' is a relative term---our search program has found $c/2$
spaceships with minimum dimension as high as 42
(Figure~\ref{fig:b27s0}) as well as narrower spaceships with
period as high as nine.

\item Sometimes small non-spaceship objects, such as puffers,
wickstretchers, or replicators, can be combined by human engineering into
a spaceship. For instance, in the near-Life rule B37/S23, the
R-pentomino pattern acts as an (unstable) puffer. 
Figure~\ref{fig:puddlejump} depicts a $9c/28$ spaceship in which a row of
six pentominoes stabilize each other while leaving behind a trail of
still lifes, which are cleaned up by a second row of four pentominoes.
Occasionally, spaceships found by our search software will
appear to have this sort of structure (Figure~\ref{fig:b27s0}).
The argument for the existence of Life spaceships with any rational
slope would also lead to patterns of this type. Several such spaceships
have been constructed by Dean Hickerson, notably the $c/12$ diagonal
``Cordership'' in Conway's Life, discovered by him in 1991.
Hickerson's web page
\url{http://math.ucdavis.edu/~dean/RLE/slowships.html}
has more examples of structured ships.

\item The remaining spaceships have large size, large period, and little
internal structure.  We believe such spaceships should exist, but none
are known and we know of no effective method for finding them.
\end{itemize}

\section{State Space}
\label{sec:state}

\begin{figure}[t]
$$\includegraphics{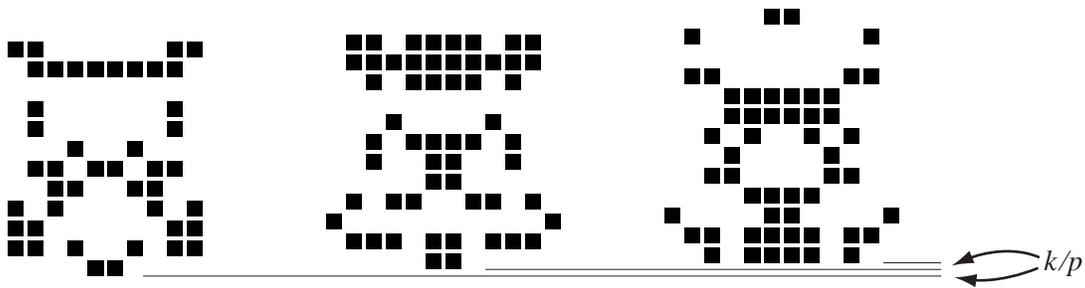}$$
\caption{The three phases of the $c/3$ ``turtle'' Life spaceship,
shifted by $1/3$ cell per phase.}
\label{fig:turtle}
\end{figure}

\begin{figure}[t]
$$\includegraphics{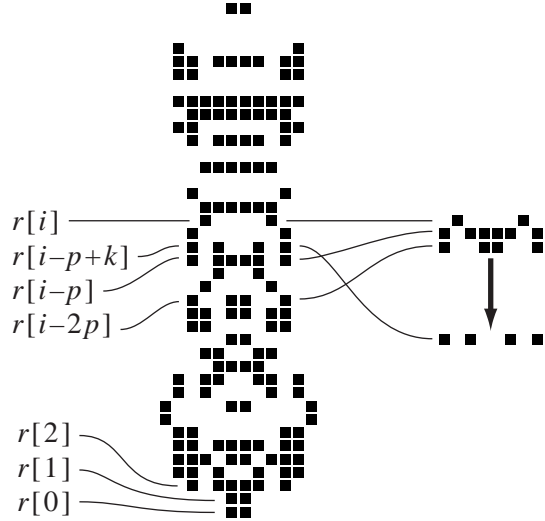}$$
\caption{Merged sequence of rows from all three phases of the turtle,
illustrating equation ($*$).}
\label{fig:merged}
\end{figure}

Due to the way our search is structured, we need
to arrange the rows from all phases of the pattern we are searching for
into a single sequence.  We now show how to do this in a way that falls
out naturally from the motion of the spaceship.

Suppose we are searching for a spaceship that moves $k$ units down every
$p$ generations.  For simplicity of exposition, we will assume
that $\gcd(k,p)=1$. We can then think of the spaceship we are searching
for as living in a cellular automaton modified by shifting the grid
upward $k/p$ units per generation.
In this modified automaton, the shifting of the grid exactly offsets the
motion of the spaceship, so the ship acts like an oscillator instead of
like a moving pattern.  We illustrate this shifted grid with
the turtle, a $c/3$ spaceship in Conway's Life
(Figure~\ref{fig:turtle}).

Because of the shifted grid, and the assumption that $\gcd(k,p)=1$, each
row of each phase of the pattern exists at a distinct vertical position.
We form a  doubly-infinite sequence
$$\ldots r[-2], r[-1], r[0], r[1], r[2] \ldots$$
of rows, by taking each row from each phase in order by the rows' vertical
positions (Figure~\ref{fig:merged}). For any $i$, the three rows
$r[i-2p],r[i-p],r[i]$ form a contiguous height-three strip in a single
phase of the pattern, and we can apply the cellular automaton rule
within this strip to calculate
\begin{equation}
r[i-p+k]=\evolve(r[i-2p],r[i-p],r[i]).
\tag{$*$}
\end{equation}
Conversely, any doubly-infinite sequence of rows, in which equation ($*$)
is satisfied for all $i$, and in which there are only finitely many live
cells, corresponds to a spaceship or sequence of spaceships. Further, any
finite sequence of rows can be extended to such a doubly infinite sequence
if the first and last $2p$ rows of the finite sequence contain only dead
cells.

Diagonal spaceships, such as Life's glider, can be handled in this
framework by modifying equation ($*$) to shift rows $i$, $i-p$, and $i-2p$
with respect to each other before performing the evolution rule.
We can also handle glide-reflect spaceships such as
Life's small $c/2$ spaceships, by modifying equation ($*$) to
reverse the order of the cells in
row $i-p+k$ (when $k$ is odd) or in the two
rows $i-p$ and $i-p+k$ (when $p$ is odd).  Note that, by the assumption
that $\gcd(k,p)=1$, at least one of $k$ and $p$ will be odd. In
these cases, $p$ should be considered as the half-period of the spaceship,
the generation at which a flipped copy of the original pattern appears.
Searches for which $\gcd(k,p)>1$ can be handled by adjusting the indices
in equation ($*$) depending on the phase to which row $r[i]$ belongs.

Our state space, then, consists of finite sequences of rows,
such that equation ($*$) is satisfied whenever all four rows in the
equation belong to the finite sequence.
The initial state for our search will be a sequence of $2p$ rows,
all of which contain only dead cells.
If our search discovers another state in which the last $2p$ rows also
contain only dead cells, it outputs the pattern formed by
every $p$th row of the state as a spaceship.

As in many game playing programs, we use a {\em transposition table} to
detect equivalent states, avoid repeatedly searching the same states, and
stop searching in a finite amount of time even when the state space may be
infinite. We would like to define two states as being equivalent if,
whenever one of them can be completed to form a spaceship, the same
completion works for the other state as well. However, this notion of
equivalence seems too difficult to compute (it would require us to be
able to detect states that can be completed to spaceships, but if we
could do that then much of our search could be avoided). So, we use a
simpler sufficient condition: two states are equivalent if their last
$2p$ rows are identical.  If our transposition table detects two
equivalent states, the longer of the two is eliminated, since it can not
possibly lead to the shortest spaceship for that rule and period.

We can form a finite directed graph, the {\em de Bruijn graph}
\cite{McI-still-88,McI-zoo-88}, by forming a vertex for each equivalence
class of states in our state space, and an edge between two vertices
whenever a state in one equivalence class can be extended to form a state
in the other class. The size of the de Bruijn graph provides a rough
guide to the complexity of a spaceship search problem.
If we are searching for patterns with width $w$, the number of vertices
in this de Bruijn graph is $2^{2pw}$.  For instance, in the search for
the weekender spaceship, the effective width was nine (due to an
assumption of bilateral symmetry), so the de Bruijn graph contained
$2^{126}$ vertices.  Fortunately, most of the vertices in this graph
were unreachable from the start state.

Coe's search program {\tt knight} uses a similar state space formed by
sequences of pattern rows, but only uses the rows from a single phase of
the spaceship. In place of our equation ($*$), he evolves subsequences of
$2p+1$ rows for $p$ generations and tests that the middle row of the
result matches the corresponding row of the subsequence.  As with our
state space, one can form a (different) de Bruijn graph by forming
equivalence classes according to the last $2p$ rows of a state.  Coe's
approach has some advantages; for instance, it can find patterns in which
some phases exceed the basic search width (as occurred in Coe's $c/5$
spaceship). However, it does not seem to allow the fast neighbor-finding
techniques we describe in Section~\ref{sec:nbr}.

\section{Search Strategies}

There are many standard algorithms for searching state spaces
\cite{Zha-99}, however each has some drawbacks in our application:

\begin{itemize}
\item Depth first search requires setting an arbitrary depth limit to
avoid infinite recursion.  The patterns it finds may be much
longer than necessary, and the search may spend a long time exploring
deep regions of the state space before reaching a spaceship.
Further, DFS does not make effective use of the large amounts of memory
available on modern computers.

\item Breadth first search is very effective for small searches, but
quickly runs out of memory for larger searches, even
when large amounts of memory are available.

\item Depth first iterative deepening \cite{Kor-AI-85} has been proposed
as a method of achieving the fast search times of breadth first search
within limited space.  However, our state space often does not have the
exponential growth required for iterative deepening to be
efficient; rather, as the search progresses from level to level the
number of states in the search frontier
can fluctuate up and down, and typically has a particularly large bulge in
the earlier levels of the search.  The overall depth of the search (and
hence the number of deepening iterations) can often be as large as several
hundred.  For these reasons, iterative deepening can be much slower than
breadth first search.  Further, the transposition table used to detect
equivalent states does not work as well with depth first as with breadth
first search: to save space, we represent this table as a collection of
pointers to states, rather than explicitly listing the $2p$ rows needed to
determine equivalence, so when searching depth first we can only detect
repetitions within the current search path. Finally, depth first search
does not give us much information about the speed at which the search is
progressing, which we can use to narrow the row width when the search
becomes too slow.

\item Other techniques such as the A$^*$ algorithm, recursive best first
search, and space-bounded best first search, depend on information
unavailable in our problem, such as varying edge weights or admissable
estimates of the distance to a solution.
\end{itemize}

Therefore, we developed a new search algorithm that combines the
best features of breadth first and iterative deepening search,
and that takes advantage of the fact that, in our search problem, many
branches of the search eventually lead only to dead ends.
Our method resembles the MREC algorithm of Sen and Bagchi
\cite{SenBag-IJCAI-89}, in that we perform deepening searches from the
breadth first search frontier, however unlike MREC we use the deepening
stages to prune the search tree, allowing additional breadth first
searching.

Our search algorithm begins by performing a standard breadth first
search.  We represent each state as a single row together with a
pointer to its predecessor, so the search must maintain the entire
breadth first search tree. By default, we allocate storage for $2^{22}$
nodes in the tree, which is adequate for small searches yet well within
the memory limitations of most computers.

On larger searches (longer than a minute or so), the breadth first
search will eventually run out of space.  When this happens, we
suspend the breadth first search and perform a round of depth first
search, starting at each node of the current breadth first search queue. 
This depth first search has a depth limit which is normally $\delta$
levels beyond the current search frontier, for a small value $\delta$
that we set in our implementation to equal the period of the pattern we
are searching for. (Setting $\delta$ to a fixed small constant would
likely work as well.)  However, if a previous round of depth first
searching reached a level past the current search frontier, we instead
limit the new depth first search round to $\delta$ levels beyond the
previous round's limit.

When we perform a depth first search from a BFS queue node,
one of three things can happen.  First, we may discover a spaceship; in
that case we terminate the entire search.  Second, we may reach the
depth limit; in that case we terminate the depth first
search and move on to the next BFS queue node.  Third, the depth first
search may finish normally, without finding any spaceships or deep nodes.
In this case, we know that the root of the search leads only to dead
ends, and we remove it from the breadth first search queue.

After we have performed this depth first search for all nodes of the
queue, we compact the remaining nodes and continue with the previously
suspended breadth first search.  Generally, only a small fraction of the
previous breadth first search tree remains after the compaction, leaving
plenty of space for the next round of breadth first search.

There are two common modes of behavior for this searching algorithm,
depending on how many levels of breadth first searching occur between
successive depth first rounds.  If more than $\delta$ levels occur
between each depth first search round, then each depth first search is
limited to only $\delta$ levels beyond the breadth first frontier, and the
sets of nodes searched by successive depth first rounds are completely
disjoint from each other.  In this case, each node is searched at most
twice (once by the breadth first and once by the depth first parts of our
search), so we only incur a constant factor slowdown over the more
memory-intensive pure breadth first search.  In the second mode of
behavior, successive depth first search rounds begin from frontiers that
are fewer than
$\delta$ levels apart.  If this happens, the $i$th round of depth first
search will be limited to depth $i\cdot\delta$, so the search resembles a
form of iterative deepening.  Unlike iterative deepening, however, the
breadth first frontier always makes some progress, permanently removing
nodes from the actively searched part of the state space. Further, the
early termination of depth-first searches when they reach deep nodes
allows our algorithm to avoid searching large portions of the state space
that pure iterative deepening would have to examine.

On typical large searches, the amount of deepening can be comparable to
the level of the breadth-first search frontier. For instance, in the
search for the weekender, the final depth first search round occurred when
the frontier had reached level 90, and this round searched from each
frontier node to an additional 97 levels.  We allow the user to supply a
maximum value for the deepening amount; if this maximum is reached, the
state space is pruned by reducing the row width by one cell and the
depth first search limit reverts back to $\delta$.

There is some possibility that a spaceship
found in one of the depth first searches may be longer than the optimum,
but this has not been a problem in practice. Even this small amount of
suboptimality could be averted by moving on to the next node instead of
terminating the search when the depth first phase of the algorithm
discovers a spaceship.

\section{Lookahead}

Suppose that our search reaches state
$$S=r[0], r[1], \ldots r[i-1].$$
The natural set of neighboring states to consider
would be all sequences of rows $$r[0], r[1], \ldots r[i-1], r[i]$$
where the first $i$ rows match state $S$
and we try all choices of row $r[i]$ that satisfy
equation~($*$).

However, it is likely that some of these choices will result in
inconsistent states for which equation ($*$) can not be satisfied the
next time the new row $r[i]$ is involved in the equation.  Since that
next time will not occur until we choose row $r[i+p-k]$, any work
performed in the intermediate levels of the search between these two rows
could be wasted. To avoid this problem, when making choices for row
$r[i]$, we simultaneously search for pairs of rows $r[i]$ and $r[i+p-k]$
satisfying both equation ($*$) and its shifted form
\begin{equation}
r[i]=\evolve(r[i-p-k],r[i-k],r[i+p-k]).
\tag{L}
\end{equation}
Note that $r[i-p-k]$ and $[i-k]$ are both already present in state
$S$ and so do not need to be searched for.
We use as the set of successors to row $S$
the sequences of rows
$r[0], r[1], \ldots r[i-1], r[i]$
such that equation ($*$) is true and equation (L) has a solution.

One could extend this idea further, and (as well as searching for rows
$r[i]$ and $r[i+p-k]$) search for two additional rows
$r[i+p-2k]$ and $r[i+2p-2k]$ such that the double lookahead equations
\begin{equation}
\arraycolsep=0pt
\begin{array}{rcl}
r[i-k] &{}={}& \evolve(r[i-p-2k],r[i-2k],r[i+p-2k])\hbox{\quad and}\\
r[i+p-k] &{}={}& \evolve(r[i-2k],r[i+p-2k],r[i+2p-2k])
\end{array}
\tag{LL}
\end{equation}
have a solution. However this double lookahead technique would greatly
increase the cost of searching for the successor states of $S$ and
provide only diminished returns.  In our implementation, we use a much
cheaper approximation to this technique: for every three consecutive
cells of the row
$r[i+p-k]$ being searched for as part of our single lookahead
technique, we test that there could exist five consecutive cells in
rows $r[i+p-2k]$ and $r[i+2p-2k]$ satisfying equations (LL) for those
three cells.  The advantage of this test over the full search for rows
$r[i+p-2k]$ and $r[i+2p-2k]$ is that it can be performed with simple
table lookup techniques, described in the next section.

\begin{figure}[t]
$$\includegraphics{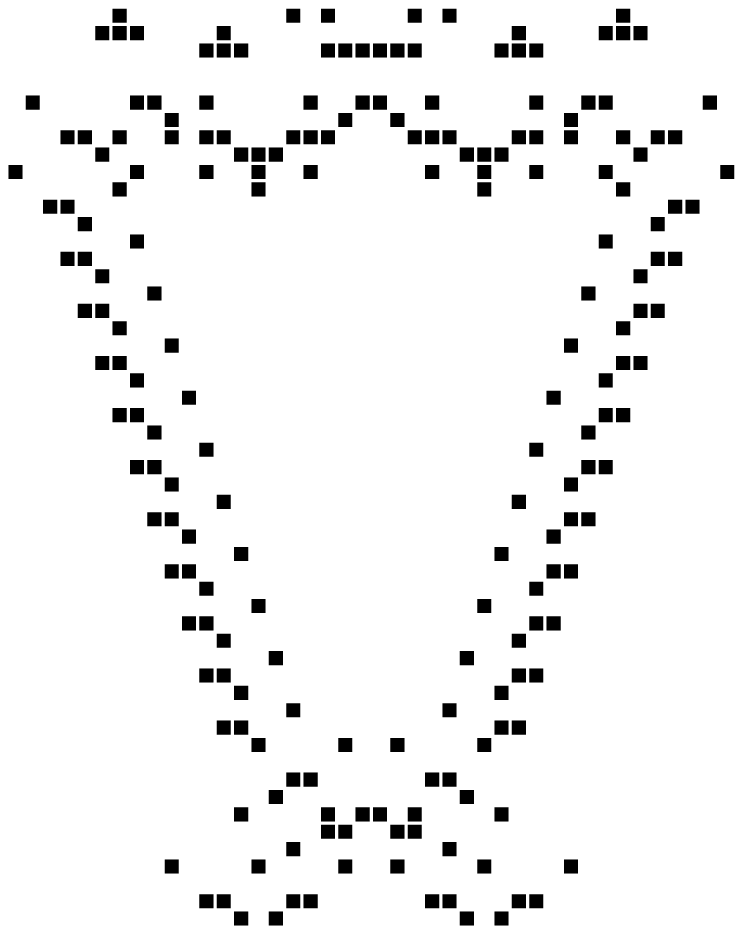}$$
\caption{$42\times 53$ $c/2$ spaceship in rule B27/S0.}
\label{fig:b27s0}
\end{figure}

For the special case $p=2$, the double lookahead technique described
above does not work as well, because $r[i+p-2k]=r[i]$.
Instead, we perform a reachability computation on a de Bruijn graph
similar to the one used for our state space, five cells wide with
don't-care boundary conditions, and test whether triples of consecutive
cells from the new rows $r[i]$ and $r[i+p-k]$ correspond to a de Bruijn
graph vertex that can reach a terminal state.
This technique varies considerably in effectiveness, depending on the
rule: for Life, 18.5\% of the 65536 possible patterns are pruned as
being unable to reach a terminal state, and for B27/S0
(Figure~\ref{fig:b27s0}) the number is 68.3\%,  but for many rules it can
be 0\%.

\section{Fast Neighbor-Finding Algorithm}
\label{sec:nbr}

To complete our search algorithm, we need to describe how to find the
successors of each state. That is, given a state
$$S=r[0], r[1], \ldots r[i-1]$$
we wish to find all possible rows $r[i]$
such that (1) $r[i]$ satisfies equation ($*$),
(2) there exists another row $r[i+p-k]$ such that
rows $r[i]$ and $r[i+p-k]$ satisfy equation (L), and
(3) every three consecutive cells of $r[i+p-k]$
can be part of a solution to equations (LL).

Note that (because of our approximation to equations (LL))
all these constraints involve only triples
of adjacent cells in unknown rows.  That is, equation ($*$) can be phrased
as saying that every three consecutive cells of $r[i]$, $r[i-p]$, and
$r[i-2p]$ form a $3\times 3$ square such that the result of the evolution
rule in the center of the square is correct; and equation (L) can be
phrased similarly.  For this reason, we need a representation of the
pairs of rows $r[i],r[i+p-k]$ in which we can access these triples of
adjacent cells.

\begin{figure}[t]
$$\includegraphics{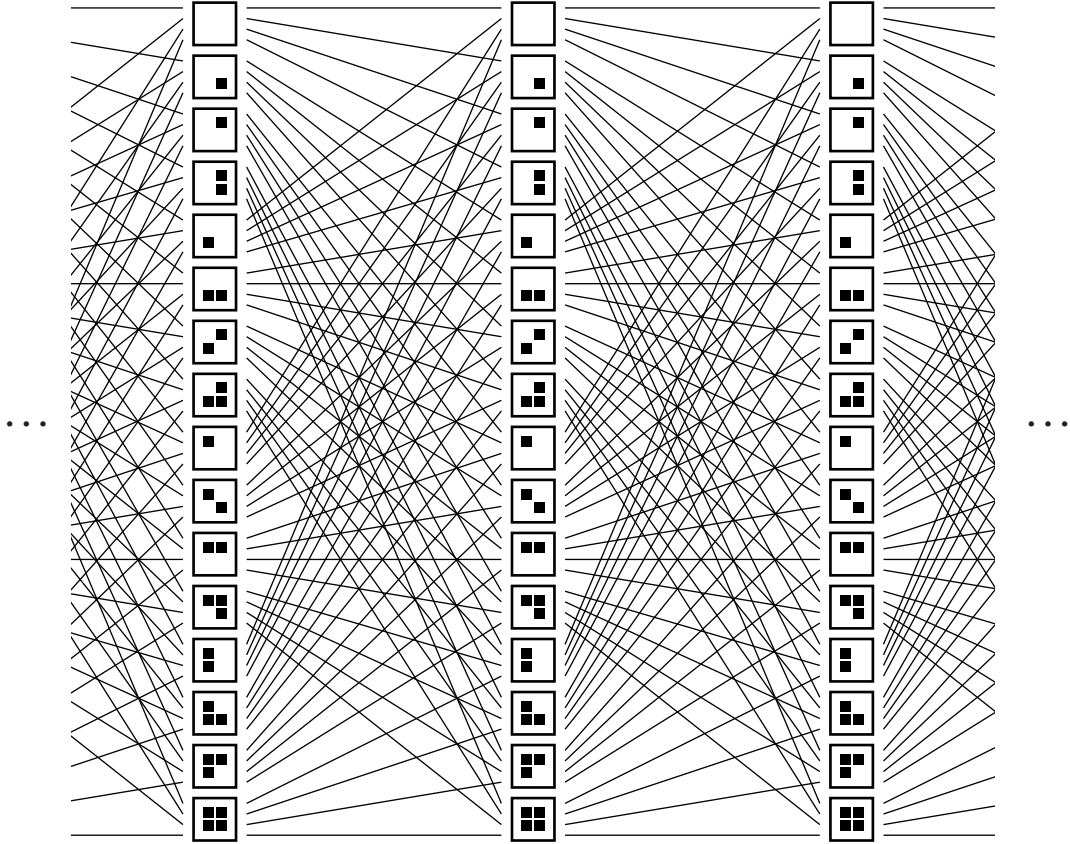}$$
\caption{Graph formed by placing each of 16 $2\times 2$ blocks of cells in
each column, and connecting a block in one column to a block in the next
column whenever the two blocks overlap to form a $2\times 3$ block.
Paths in the graph represent pairs of rows $r[i],r[i+p-k]$;
by removing edges from the graph we can constrain triples of adjacent
cells from these rows.}
\label{fig:columns}
\end{figure}

Such a representation is provided by the graph depicted in
Figure~\ref{fig:columns}.  This graph (which like our state space can be
viewed as a kind of de Bruijn graph) is formed by a sequence of columns of
vertices. Each column contains 16 vertices, representing the 16 ways of
forming a
$2\times 2$ block of cells. We connect a vertex in one column with a
vertex in the next column whenever the two blocks overlap to form a
single $2\times 3$ block of cells; that is, an edge exists whenever the
right half of the left
$2\times 2$ block matches the left half of the right $2\times 2$ block.

If we have any two rows of cells, one placed above the other,
we can form a path in this graph in which the vertices correspond
to $2\times 2$ blocks drawn from the pair of rows.
Conversely, the fact that adjacent blocks are required to match implies
that any path in this graph corresponds to a pair of rows.

Since each triple of adjacent cells from the two rows corresponds to an
edge in this graph, the constraints of equations ($*$), (L), and (LL)
can be handled simply by removing the edges from the graph that
correspond to triples not satisfying those constraints.  In this
constrained subgraph, any path corresponds to a pair of rows satisfying
all the constraints.  Thus, our problem has been reduced to finding the
constrained subgraph, and searching for paths through it between
appropriately chosen start and end terminals.  The choice of which
vertices to use as terminals depends on the symmetry type of the pattern
we are searching for.

This search can be understood as being separated into three stages,
although our actual implementation interleaves the first two of these.

In the first stage, we find the edges of the graph corresponding to
blocks of cells satisfying the given constraints.  We represent the set
of 64 possible edges between each
pair of columns in the graph (as shown in Figure~\ref{fig:columns}) as a
64-bit quantity, where a bit is nonzero if an edge exists and zero
otherwise. The set of edges corresponding to blocks satisfying equation
($*$) can be found by a table lookup with an index that encodes the
values of three consecutive cells of $r[i-p]$ and
$r[i-2p]$ together with one cell of $r[i-p+k]$. Similarly, the set of
edges corresponding to blocks satisfying equation (L) can be found by a
table lookup with an index that encodes the values of three consecutive
cells of $r[i-k]$ and
$r[i-p-k]$.  In our implementation, we combine these two constraints
into a single table lookup.  Finally, the set of edges corresponding to
blocks satisfying equations (LL) can be found by a table lookup with an
index that encodes the values of five consecutive cells of $r[i-2k]$ and
$r[i-p-2k]$ together with three consecutive cells of $r[i-k]$.
The sets coming from equations ($*$), (L), and (LL) are combined with a
simple bitwise Boolean and operation.  The various tables used by this
stage depend on the cellular automaton rule, and are precomputed based on
that rule before we do any searching.

In the second stage, we compute a 16-bit quantity for each column of
vertices, representing the set of vertices in that column that can be
reached from the start terminal.  This set can be computed from the set
of reachable vertices in the previous column by a small number of shift
and mask operations involving the sets of edges computed in the previous
stage.

In the third stage, we wish to find the actual rows $r[i]$ that
correspond to the paths in the constrained graph.  We perform a
backtracking search for these rows, starting with the end terminal of
the graph.  At each step, we maintain a 16-bit quantity,
representing the set of vertices in the current column of the graph
that could reach the end terminal by a path matching the current partial
row.  To find the possible extensions of a row, we find the predecessors
of this set of vertices in the graph by a small number of shift and mask
operations (resembling those of the previous stage) and separate these
predecessors into two subsets, those for which the appropriate cell of
$r[i]$ is alive or dead.  We then continue recursively in one or both of
these subsets, depending on whichever of the two has a nonempty
intersection with the set of vertices in the same column that can reach
the start vertex.

Because of the reachability computation in the second stage, the third
stage never reaches a dead end in which a partial row
can not be extended.  Therefore, this algorithm spends a total amount of
time at most proportional to the width of the rows (in the first two
stages) plus the width times the number of successor states (in the third
stage). The time for the overall search algorithm is
bounded by the width times the number of states reached.

\begin{figure}[t]
$$\includegraphics[width=\textwidth]{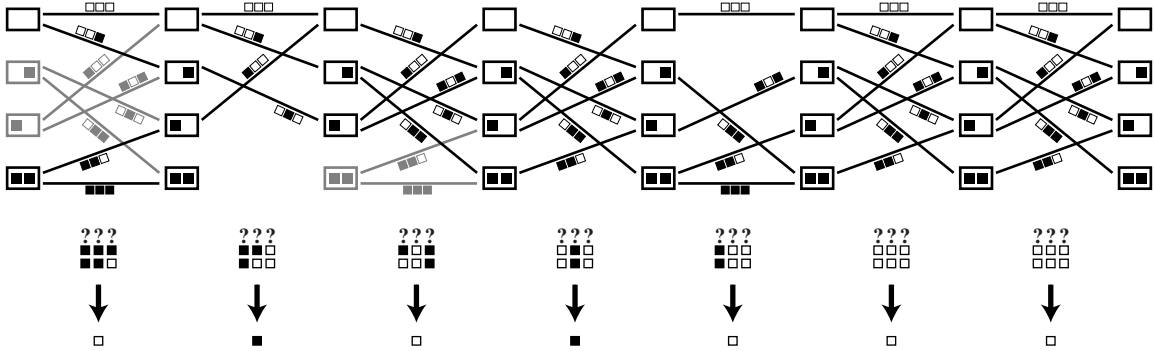}$$
\caption{Simplified example (without lookahead) of graph representing
equation ($*$) for the rows from Fig.~\ref{fig:merged}. Gray
regions represent portions of the graph not reachable from the two
start vertices.}
\label{fig:colex}
\end{figure}

A simplified example of this graph representation of our problem is
depicted in Figure~\ref{fig:colex}, which shows the graph formed by the
rows from the Life turtle pattern depicted in Figure~\ref{fig:merged}.  To
reduce the complexity of the figure, we have only incorporated equation
($*$), and not the two lookahead equations.  Therefore, the vertices in
each column represent the states of two adjacent cells in row
$r[i]$ only, instead of also involving row $r[i+p-k]$, and there are four
vertices per column instead of sixteen.  Each edge represents the state
of three adjacent cells of $r[i]$, and connects vertices corresponding
to the left two and right two of these cells; we have marked each edge
with the corresponding cell states.

Due to the symmetry of the turtle, the
effective search width is six, so we need to enforce equation ($*$)
for six cells of row
$r[i-p+k]$.  Six of the seven columns of edges in the graph correspond
to these cells; the seventh represents a cell outside the search width
which must remain blank to prevent the pattern from growing beyond its
assigned bounds.  Below each of these seven columns, we have shown the
cells from previous rows $r[i-p]$, $r[i-2p]$, and $r[i-p+k]$ which
determine the set of edges in the column and which are concatenated
together to form the table index used to look up this set of edges.

The starting vertices in this example are the top and bottom left ones
in the graph, which represent the possible states for the center two cells
of $r[i]$ that preserve the pattern's symmetry.  The destination vertex
is the upper right one; it represents the state of two cells in
row $r[i]$ beyond the given search width, so both must be blank.  Each
path from a start vertex to the destination vertex represents a possible
choice for the cells in row $r[i]$ that would lead to the correct
evolution of row $r[i-p+k]$ according to the rules of Conway's life. 
There are 13 such paths in the graph shown.  The reachability
information computed in the second stage of the algorithm is depicted by
marking unreachable vertices and edges in gray; in this example, as well
as the asymmetric states in the first column, there is one more
unreachable vertex.

For this simplified example, a list of all 13
paths in this graph could be found in the third stage by a recursive depth
first search from the destination vertex backwards, searching only the
black edges and vertices. Thus even this simplified representation
reduces the number of row states that need to be considered from 64 to 13,
and automatically selects only those states for which the evolution rule
leads to the correct outcome. The presence of lookahead complicates the
third stage in our actual program since multiple paths can correspond to
the same value of row~$r[i]$; the recursive search procedure described
above finds each such value exactly once.

\section{Conclusions}

To summarize, we have described an algorithm that finds spaceships in
outer totalistic Moore neighborhood cellular automata, by a hybrid
breadth first iterative deepening search algorithm in a state space
formed by partial sequences of pattern rows.  The algorithm represents
the successors of each state by paths in a regularly structured graph with
roughly $16w$ vertices; this graph is constructed by performing
table lookups to quickly find the sets of edges representing the
constraints of the cellular automaton evolution rule and of our lookahead
formulations.  We use this graph to find a state's successors by
performing a 16-way bit-parallel reachability algorithm in this graph,
followed by a recursive backtracking stage that uses the reachability
information to avoid dead ends.  Empirically, the algorithm works
well, and is able to find large new patterns in many cellular automaton
rules.

This work raises a number of interesting research questions, beyond the
obvious one of what further improvements are possible to our search
program:

\begin{figure}[t]
$$\includegraphics[width=\textwidth]{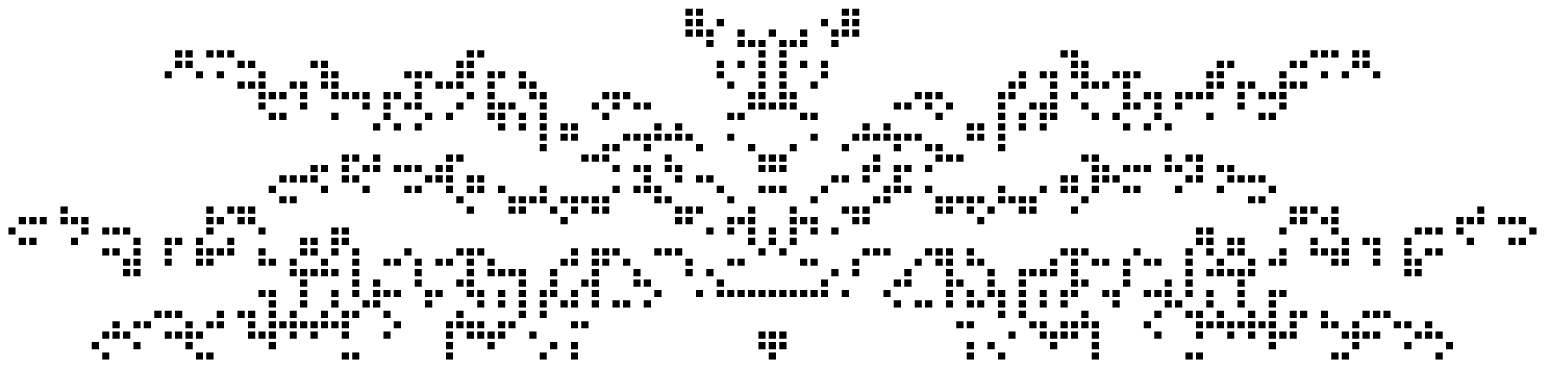}$$
\caption{Large period-114 $c/3$ blinker puffer in Life,
found by David Bell, Jason Summers, and Stephen Silver using a
combination of automatic and human-guided searching.}
\label{fig:p114}
\end{figure}

% message from Silver 7 Jan 2000

\begin{itemize}
\item The algorithm described here, and the two search algorithms
previously used by Hickerson and Coe, use three different state spaces.
Do these spaces really lead to different asymptotic search performance,
or are they merely three different ways of looking at the same thing? Can
one make any theoretical arguments for why one space might work better
than another?

\item Is it possible to explain the observed fluctuations in size of the
levels of the state space?  A hint of an explanation for the fact that
fluctuations exist comes from the idea that we typically run as narrow as
possible a search as we can to find a given period spaceship.
Increasing the width seems to increase the branching factor, and
perhaps we should expect to find spaceships as soon as the state space
develops infinite branches, at which point the branching factor will
likely be quite close to one.  However this rough idea depends on the
unexplained assumption that the start of a spaceship is harder to find
than the tail, and it does not explain other features of the state space
size such as a large bulge near the early levels of the search.

\item We have mentioned in Section~\ref{sec:state} that one can use the
size of the de Bruijn graph as a rough guide to the complexity of the
search. However, our searches typically examine far fewer nodes than are
present in the de Bruijn graph.  Further, there seem to be other factors
that can influence the search complexity; for instance, increasing $k$
seems to decrease the running time, so that e.g. a width-nine search for
a $c/7$ ship in Life would likely take much more time than the search for
the weekender.  Can we find a better formula for predicting search run
time?

\item What if anything can one say about the computational complexity of
spaceship searching?  Is the problem of determining whether a given outer
totalistic rule has a spaceship of a given speed or period even decidable?

\item David Bell and others have had success finding large spaceships
with ``arms'' (Figure~\ref{fig:p114}) by examining partial results from
an automatic search, placing ``don't care'' cells at appropriate
connection points, and then doing secondary searches for arms that
can complete the pattern from each connection point.  To what extent can
this human-guided search procedure be automated?

\item What other types of patterns can be found by our search techniques?
For instance, one possibility would be a search for predecessors of a
given pattern. The rows of each predecessor satisfy a consistency
condition similar to equation ($*$), and it would not be difficult to
incorporate a lookahead technique similar to the one we use for spaceship
searching. However due to the fixed depth of the search it seems that
depth first search would be a more appropriate strategy than the breadth
first techniques we are using in our spaceship searches.

\item Carter Bays has had some success using brute force
methods to find small spaceships in various three-dimensional rules
\cite{Bay-CS-87,Bay-CS-90,Bay-CS-94a}, and rules on the triangular planar
lattice \cite{Bay-CS-94b}.  How well can other search methods such as the
ones described here work for these types of automata?

\item To what areas other than cellular automata can our search
techniques be applied?  A possible candidate is in document
improvement, where Bern, Goldberg, and others \cite{BerGol-ms-99} have
been developing algorithms that look for a single high-resolution image
that best matches a given set of low-resolution samples of the same
character or image.  Our graph based techniques could be used to replace
a local optimization technique that changes a single pixel at a time,
with a technique that finds an optimal assignment to an entire row of
pixels, or to any other linear sequence such as the set of pixels around
the boundary of an object.
\end{itemize}

\section*{Acknowledgements}

Thanks go to Matthew Cook, Nick Gotts, Dean Hickerson, Harold McIntosh,
Gabriel Nivasch, and Bob Wainwright for helpful comments on drafts of this
paper; to Noam Elkies, Rich Korf, and Jason Summers for useful suggestions
regarding search algorithms; to Keith Amling, David Bell, and Tim Coe,
for making available the source codes of their spaceship searching
programs; and to Richard Schroeppel and the members of the Life mailing
list for their support and encouragement of cellular automaton research.

\bibliographystyle{abuser}
\bibliography{spaceship}
\end{document}